\documentclass[preprint,12pt]{elsarticle}
\usepackage{amssymb}
\usepackage{amsmath}
\usepackage[utf8]{inputenc}
\usepackage{textgreek}
\usepackage{graphicx}
\usepackage{newunicodechar}
\usepackage{tabularx}        % 你在用 tabularx 表格
\usepackage{threeparttable}  % 提供 tablenotes 环境
\usepackage{booktabs}        % 可选：更好看的表格线
\usepackage{float}
\usepackage{placeins}
\usepackage{chngcntr}

\newunicodechar{≤}{\leq}
\newunicodechar{≥}{\geq}

\journal{neuroimage}

\begin{document}
\begin{frontmatter}
%TC:ignore
\title{Language-Enhanced Generative Modeling for Amyloid PET Synthesis from MRI and Blood Biomarkers}

\author[1,7]{Zhengjie Zhang}
\author[2,3,7]{Xiaoxie Mao}
\author[4]{Qihao Guo}
\author[1]{Shaoting Zhang}
\author[3]{Qi Huang\corref{cor1}}
\author[5]{Mu Zhou\corref{cor1}}
\author[3]{Fang Xie\corref{cor1}}
\author[1,6]{Mianxin Liu\corref{cor1}}

\fntext[7]{These authors contributed equally to this work.}

\cortext[cor1]{%
\begin{tabular}{@{}l@{}}
\textsuperscript{} Corresponding authors: liumianxin@pjlab.org.cn (M.~Liu); fangxie@fudan.edu.cn (F.~Xie) \\
 muzhou1@gmail.com (M.~Zhou); hq\_1124@163.com (Q.~Huang)
\end{tabular}
}

\affiliation[1]{organization={Shanghai Artificial Intelligence Laboratory},
                % addressline={No.700, Changyang Road},
                city={Shanghai},
                postcode={200082},
                country={China}}

\affiliation[2]{organization={School of Medicine, Xiamen University},
                city={Xiamen},
                state={Fujian},
                country={China}}

\affiliation[3]{organization={Department of Nuclear Medicine \& PET Center, Huashan Hospital, Fudan University},
                city={Shanghai},
                country={China}}

\affiliation[4]{organization={Department of Gerontology, Shanghai Jiao Tong University Affiliated Sixth People’s Hospital},
                city={Shanghai},
                country={China}}

\affiliation[5]{organization={Department of Computer Science, Rutgers University},
                city={New Brunswick},
                state={New Jersey},
                country={United States}}

\affiliation[6]{organization={Shenzhen Institutes of Advanced Technology, Chinese Academy of Sciences},
                city={Shenzhen},
                country={China}}

%• Abstract 
\begin{abstract}
Alzheimer’s disease (AD) diagnosis heavily relies on amyloid-beta positron emission tomography (Aβ-PET), which is limited by its high cost and limited accessibility. This study explores whether Aβ-PET spatial patterns can be predicted from blood-based biomarkers (BBMs) and magnetic resonance imaging (MRI) scans. We collected Aβ-PET images, T1-weighted MRI scans, and BBMs from 566 participants. A language-enhanced generative model, driven by a large language model (LLM) and multimodal information fusion, was developed to synthesize PET images. Synthesized images were evaluated for image quality, diagnostic consistency, and clinical applicability within a fully automated diagnostic pipeline. The synthetic PET images closely resemble real PET scans in both structural details (SSIM = 0.920 ± 0.003) and regional patterns (Pearson’s R = 0.955 ± 0.007). In physician evaluations, the diagnostic outcomes using synthetic PET show high agreement with real PET–based diagnoses (accuracy = 0.80). We developed a fully automatic AD diagnostic pipeline integrating PET synthesis and classification. The synthetic PET–based model (AUC = 0.78) outperforms T1-based (AUC = 0.68) and BBM-based (AUC = 0.73) models, while combining synthetic PET and BBMs further improved performance (AUC = 0.79). Our method significantly outperform conventional MRI- and BBM-based synthesis and diagnosis methods. Ablation analysis supports the advantages of LLM integration and prompt engineering. Our language-enhanced generative model demonstrates strong capability in synthesizing realistic PET images, enhancing the utility of MRI and BBMs for Aβ spatial pattern assessment. The developed fully automatic and cost-effective AD diagnostic pipeline improves the diagnostic workflow for Alzheimer’s disease.
\end{abstract}

%% Keywords

\end{frontmatter}

\section{Introduction}
\label{sec:intro}

Alzheimer’s disease (AD) is a neurodegenerative disorder causing progressive cognitive and functional decline in patients \cite{scheltens2021alzheimer, masters2015alzheimer}. Although treatments such as lecanemab and donanemab have shown promise, AD progression remains irreversible \cite{veitch2019understanding, van2023lecanemab, srivastava2021alzheimer}. Therefore, timely and accurate diagnosis is essential for effective AD screening and management. To date, amyloid-beta (Aβ) and tau pathology detection primarily rely on positron emission tomography (PET) imaging and cerebrospinal fluid sampling \cite{jack2024revised, van2022atn, jack2018nia}. In particular, PET imaging remains the only in vivo, non-invasive method capable of providing refined spatial distributions of Aβ and tau depositions, which are crucial for AD subtyping, staging, and prognosis \cite{biel2021tau, mattsson2019staging, ossenkoppele2021accuracy}. However, PET imaging is limited by its high cost, low accessibility, and the potential radiation exposure to patients \cite{rice2017diagnostic, nordberg2010use}.

To overcome these limitations, generative models have emerged as a promising approach to synthesise realistic PET images from more cost-effective, accessible, and non-radiative data modalities \cite{hu2021bidirectional, zhang2022bpgan}. Previous studies have explored Aβ-PET synthesis using various data sources, including T1-weighted MRI (T1 images) and functional MRI (fMRI) \cite{vega2024image, li2023individualized}. T1 images provide high-resolution anatomical details, while fMRI captures dynamic brain activity and functional connectivity. Although both modalities offer clinically relevant information, they are not directly coupled with Aβ deposition \cite{jack2010hypothetical, hasani2021systematic}. Consequently, analytical methods based on these modalities often struggle to achieve high-quality, reliable Aβ-PET synthesis.

Recent advances in blood-based biomarkers (BBMs) offer a sensitive, cost-effective assessment of Aβ pathology at the global level \cite{mielke2024alzheimer, hansson2022alzheimer, hansson2023blood, olsson2016csf, teunissen2022blood}. BBM-based methods have successfully predicted Aβ-PET positivity and global standardised uptake value ratios (SUVRs) \cite{janelidze2020plasma, barthelemy2024highly, meyer2024clinical, wisch2023predicting}. Importantly, BBMs are minimally invasive, easy to collect, and inexpensive to apply in clinical settings \cite{palmqvist2024blood, meyer2024clinical}. However, while BBMs reflect the macroscopic state of Aβ pathology, they lack spatial resolution and are insufficient to represent detailed deposition patterns captured by PET imaging. Given the complementary nature of BBMs and MRI, integrating these modalities into a multimodal framework could enable refined, high-quality, and cost-effective Aβ-PET synthesis.

While T1 images offer detailed spatial information, BBMs are numerical data without inherent spatial context. This modality disparity presents a major challenge for multimodal fusion. Leveraging the capabilities of large medical language models (LLMs) and prompt engineering \cite{zhang2024data, singhal2023large, thirunavukarasu2023large, wang2023prompt, ahmed2024med}, we introduce a language-enhanced encoder that aligns BBMs with deep clinical knowledge beyond raw numerical representation. Trained on large-scale medical corpora, the LLM captures clinically meaningful semantics. By translating non-imaging data into context-rich representations, we explore prompt-guided LLMs for extracting knowledge-enhanced BBM features. Integrating these with T1-derived spatial features, we enable more effective multimodal generative modelling for AD assessment.

In this study, we propose a language-enhanced generative model based on clinical T1 images and BBMs to synthesise realistic PET images for improved Aβ spatial pattern assessment. We comprehensively evaluate the synthetic PET images in terms of image quality, diagnostic feature consistency (including physician-based assessment), and clinical applicability using a large cohort (N = 566). The proposed model outperforms conventional methods across all evaluation dimensions, primarily due to its LLM-driven feature enhancement and prompt-based learning strategy. Our results highlight that this generative AI framework provides a cost-effective solution to support large-scale AD screening, diagnosis, and management.

\section{METHODS}\label{sec2}
\subsection{Participants and materials}
\subsubsection{Participants}

Participants for this study were recruited from Huashan Hospital and the Sixth People’s Hospital affiliated with Shanghai Jiao Tong University. The research was approved by the institutional review boards of both hospitals and all participants provided written informed consent. Data collection was registered under the identifier "ChiCTR2000036842" with the title "Construction of a Pre-clinical AD Neuroimaging Cohort Using the ATN System" (http://www.chictr.org.cn/showproj.aspx?proj=59802).

Participants were selected according to the following exclusion criteria: 1) fewer than 5 years of education, 2) a history of neurological or psychiatric disorders, 3) other neurological conditions beyond AD spectrum disorders, and 4) significant alcohol or drug abuse. In total, 566 participants were enrolled and data collection commenced in December 2018 and concluded in July 2022. The dataset includes demographic details (age, gender, years of education), MR and PET imaging data, 7 types of BBMs, and results from 8 neuropsychological tests.

\subsubsection{MR and PET imaging}

T1 images were acquired using a 3T Prisma MRI scanner (Siemens, Erlangen, Germany) at the Shanghai Jiao Tong University Affiliated Sixth People's Hospital. The imaging was performed with a 3D magnetization-prepared rapid gradient-echo sequence, employing the following parameters: repetition time = 3000 ms, echo time = 2.56 ms, flip angle = 7°, acquisition matrix = 320 × 320, in-plane resolution = 0.8 × 0.8 mm², slice thickness = 0.8 mm, and 208 sagittal slices.

Aβ-PET imaging was performed using a Biograph mCT Flow PET/CT scanner (Siemens Healthineers, Erlangen, Germany) at the Department of Nuclear Medicine \& PET Center, Huashan Hospital, Fudan University, within two weeks following the MRI scans \cite{ren2023brain}. Participants received an intravenous injection of 18F-AV45 at approximately 7.4 MBq/kg. After a 50-minute rest period, they underwent low-dose CT scanning and PET scanning. The PET scans were subsequently processed using a filtered back-projection reconstruction algorithm, which included corrections for decay, normalization, dead time, photon attenuation (based on CT), scatter, and random coincidences. The final PET images had a matrix size of 168 × 168 × 148 voxels and a voxel resolution of 2.04 × 2.04 × 1.5 mm³.

\subsubsection{Blood-based biomarkers for AD}

We collected plasma samples before PET scans, stored them at -80°C, and subsequently analysed these samples using the Quanterix Simoa HD-1 platform \cite{olsson2016csf,janelidze2020plasma}. The measurement of plasma AD biomarkers included Aβ42, Aβ40, T-tau (Neurology 3-Plex A assay kit, lot 502838), p-tau181 (Assay Kit V2, lot 502923), and neurofilament light chain (NFL, NF-light assay kit, lot 202700)\cite{pan2022non,chu2024associations}. The Aβ42/40 ratio was calculated as a normalized measure of Aβ pathology. The p-tau181/Aβ42 ratio was computed to assess the relationship between tau phosphorylation and amyloid pathology. In addition, NFL concentrations were measured to indicate the level of neurodegeneration. Technicians blinded to clinical imaging data performed the measurements to avoid potential bias. Plasma biomarker concentrations were reported in pg/mL.

\subsubsection{Neuropsychological assessment}

Systematic neuropsychological assessments (NAs) were conducted by trained neuropsychologists at Shanghai Jiao Tong University Affiliated Sixth People's Hospital. This evaluation comprised 8 assessments: Mini-Mental State Examination (MMSE), Montreal Cognitive Assessment-Basic (MoCA-B), auditory verbal learning test 30-minute long-delayed free recall (AVLT-LDR), AVLT-recognition (AVLT-R), animal fluency test (AFT), 30-item Boston naming test (BNT), shape trails test A and B (STT-A and STT-B) \cite{guo2009comparison,zhao2013clustering,zhao2013shape,zhao2015auditory,ding2016progression}. We obtained 8 neuropsychological scores from these assessments: MMSE and MoCA-B for global cognitive functions; AVLT-LDR and AVLT-R for memory function; AFT and BNT (total scores) for language function; and STT-A and STT-B (time to completion) for executive function. Higher scores indicate better cognitive ability for all measures except STT-A and STT-B, where shorter completion times reflect better executive function. It should be noted that we have included all available scores to patients. In particular, MMSE and MoCA-B scores were completed by all patients.

The criteria for diagnosing AD dementia (ADD) and mild cognitive impairment (MCI) were derived from the 2011 National Institute on Aging-Alzheimer's Association (NIA-AA) guidelines for ADD, along with a modified version of Jak and Bondi's criteria\cite{mckhann2011diagnosis, chen2024metabolic}. MCI was identified in participants who showed impairment in one or more cognitive domains or had scores more than one standard deviation below the norm in each of the three cognitive domains\cite{huang2020conceptual}. Participants exhibiting no signs of cognitive impairment were classified as cognitively unimpaired (CU).

\subsubsection{Data preprocessing}

T1 and PET images were preprocessed using Advanced Normalization Tools (ANTs, https://antspy.readthedocs.io/en/latest/index.html). For T1 images, N4 bias correction was applied to mitigate bias field effects, followed by skull stripping with a pretrained deep learning method from ANTsPyNet (https://antspy.readthedocs.io/en/latest/segmentation.html). PET images were registered to the corresponding T1 images using an affine transformation computed by ANTs after acquisition and reconstruction. The same brain mask from the T1 images was used to perform skull stripping on the PET images. To convert PET intensity values to SUVR, we normalized them against the average regional value of the bilateral cerebellar crus. These regions were identified using an affine registration from the automated anatomical labeling-116 (AAL-116) atlas, aligning the template T1 brain images to the individual T1 space. All images were then resized to 128×128×128 voxels and scaled to intensity values between 0 and 1 before being fed into the AI models.

\subsection{Language-enhanced generative framework}

\begin{figure}[H]
\centerline{\includegraphics[width=1\textwidth]{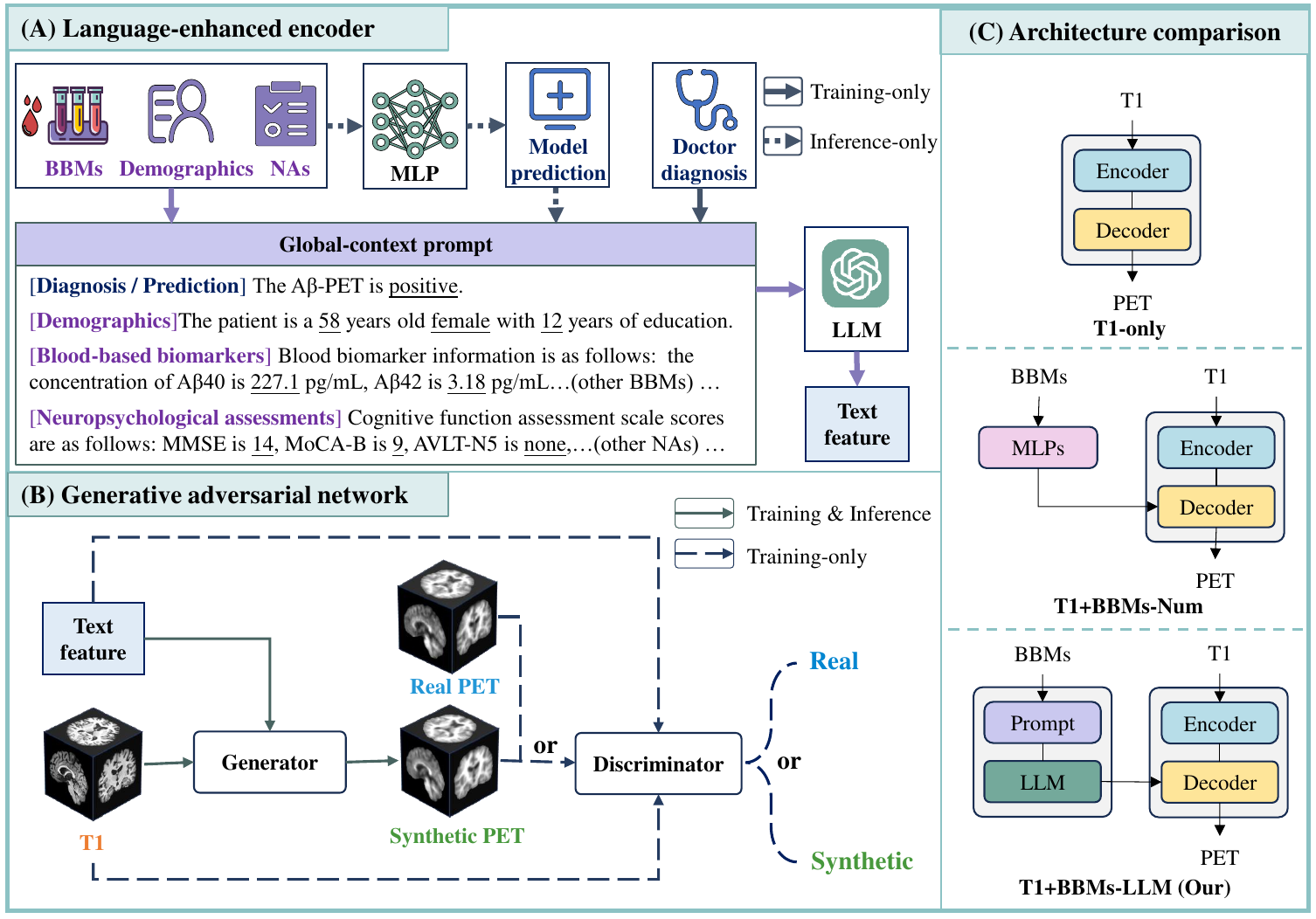}}
\caption{Overview of our proposed language-enhanced generative framework for PET synthesis. \textbf{(A) Language-enhanced encoder:} Clinical variables including demographics, BBMs, and NAs are formatted into a global-context prompt, which begins with a global PET characteristics statement (ground-truth diagnosis for training and clinical-variable-based prediction for inference, see Methods), followed by structured clinical details. The prompt is encoded into text features using a medical LLM. 
\textbf{(B) Generative adversarial network:} The generator uses T1 images along with text features to synthesize Aβ-PET images. The discriminator evaluates the authenticity of PET images by comparing them with paired T1 images and text features only during training stage. 
\textbf{(C) Architecture comparison:} This component illustrates the key differences between our method and two representative baselines. Our approach ("T1+BBMs-LLM") synthesizes PET images using T1-weighted images and BBMs encoded as text features via an LLM. In contrast, the "T1-only" baseline uses only T1 images, while the "T1+BBMs-Num" baseline incorporates BBMs as normalized numerical features. 
}
\label{fig1a}
\end{figure}

Figure \ref{fig1a} illustrates our language-enhanced generative framework for synthesizing PET images from T1 images and non-imaging data composed of patient demographics, BBMs, and NAs. The framework comprises two main components: a language-enhanced encoder and a generative adversarial network (GAN). Specifically, the language-enhanced encoder first transforms all non-imaging data into a Global-context Prompt, which summarizes key clinical information, and then encodes this prompt into semantic features using an LLM. The generative adversarial network consists of a generator and a discriminator. The generator takes T1 images and the encoded text features as inputs to produce the corresponding PET images. Meanwhile, the discriminator distinguishes between real and synthetic PET images for adversarial training. To enable effective multimodal integration, two separate text-image fusion modules are respectively designed for the generator and the discriminator.

\subsubsection{Language-enhanced encoder}

The language-enhanced encoder is designed to extract semantic features from a range of clinical variables (BBMs, demographics, and NAs). This process consists of two main stages: (1) transforming these variables into a unified textual format, referred to as the global-context prompt, and (2) encoding this prompt into semantic features using a medical LLM.

Prompt engineering involves mapping numerical clinical variables into structured text, which is crucial for guiding the LLM to produce accurate and refined outputs \cite{sahoo2024systematic, liu2022design}. To enable this, we design a global-context prompt that combines a summary-level descriptor of Aβ pathology with a structured representation of key clinical variables (see “Global-context prompt” in Figure \ref{fig1a}). This prompt provides a semantic lead-in to orient the LLM’s interpretation while integrating all relevant inputs into a coherent, language-based format for effective context modeling.

The global-context prompt consists of two components: (i) a summary diagnostic sentence reflecting the overall Aβ pathology status (e.g., “The Aβ-PET is positive/negative.”), and (ii) a structured list of 18 clinical variables, including age, gender, years of education, Aβ40, Aβ42, T-Tau, P-Tau 181, NFL, Aβ42/40, P-Tau 181/Aβ42, MMSE, MoCA-B, AVLT-N5, AVLT-N7, AFT, BNT, STT-A, and STT-B. To ensure structural consistency, missing values are explicitly denoted as “None”.

During training, the summary sentence is derived from the gold-standard Aβ-PET diagnosis obtained through expert assessment. In the inference stage, where real PET images and expert evaluations are unavailable, we instead derive a coarse prediction for constructing this summary using a prediction model based on clinical variables. Specifically, a two-layer multilayer perceptron (MLP) is trained to predict Aβ-PET positivity based on 11 clinical variables, excluding gender and 6 NAs that exhibit substantial missingness. The model's binary prediction is then used to construct the summary sentence. Notably, this predicted summary serves only as guidance for the LLM and does not explicitly determine the final characteristics of the synthetic PET image. As demonstrated in Section 3.4 and Figure S1, the generated PET images remain capable of capturing realistic Aβ pathology patterns even when the predicted summaries contain mistakes.

For the feature extraction, we utilize a pretrained BioMedBERT model \cite{chakraborty2020biomedbert} that specializes in biomedical natural language processing. Trained on a diverse range of biomedical texts (e.g., PubMed), BioMedBERT has a broad knowledge base to encode clinical text information. The model processes text data by first tokenizing it into smaller units of words. These tokens are then analysed through multiple transformer layers, which capture the context and semantic meaning of the text inputs. This critical design results in the generation of a knowledge-enhanced feature representation of the input clinical texts. 

\begin{figure}[H]
\centerline{\includegraphics[width=1\textwidth]{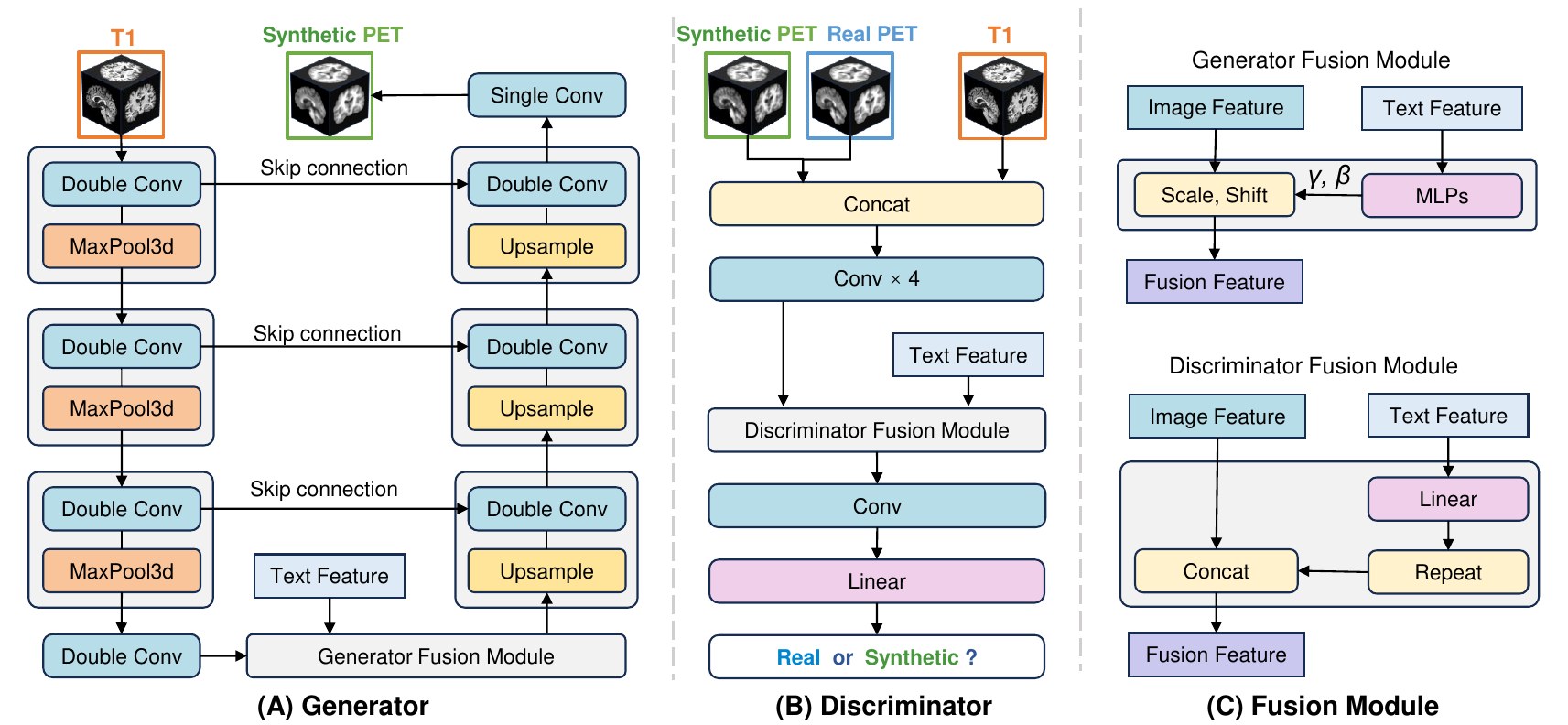}}
\caption{Detailed architecture of the proposed framework.
\textbf{(A) Generator:} The generator employs a three-layer U-Net architecture. The encoder includes three downsampling blocks with double convolutional layers and max pooling.  The encoded T1 features are then fused with text features. The decoder features three upsampling blocks with trilinear interpolation and double convolutional layers, incorporating skip connections to retain feature information. Finally, the decoder output is processed through a convolutional layer to synthesize the PET image.
\textbf{(B) Discriminator:} The discriminator begins by concatenating the T1 and PET image, followed by feature extraction using four convolutional layers. These image features are integrated with text features. The combined features are subsequently processed through a convolutional layer and a fully connected layer to produce the authenticity judgment for each PET image. 
\textbf{(C) Fusion Module:} In the generator fusion module, text features generate scale and shift parameters through non-linear multilayer perceptrons (MLPs), which adjust image features on a channel-wise basis. In the discriminator fusion module, text features are first reduced in dimensionality through a linear layer, then expanded by repeating the channels to fill the remaining dimensions, and finally concatenated with the image features.}
\label{fig1b}
\end{figure}

\subsubsection{Generator and discriminator}

In Figure \ref{fig1b}, the generator synthesizes PET images from T1 images and encoded text features, while the discriminator learns to distinguish between real and synthetic PET images. Through iterative optimization of both components, we find that the model can progressively improve the quality of the synthetic PET images.

The generator is built using a U-Net architecture, a type of convolutional neural network (CNN) \cite{cciccek20163d}. U-Net features a contracting path (encoder) that captures context through downsampling and feature compression, and a symmetric expanding path (decoder) that enables precise localization through upsampling and feature decompression. Both the encoder and decoder use convolutional layers, with skip connections established between corresponding spatial resolutions. These skip connections transmit features directly to the decoder, preserving finer details in the synthetic PET images. Additionally, the bottleneck layer of the U-Net includes a generator fusion module to integrate T1 features with text features.

The discriminator is a CNN model designed to classify PET images as either synthetic or real classes. We extend the concept of paired adversarial loss \cite{mirza2014conditional} to be implemented with three modalities. Along with using synthetic or real PET images as inputs, the discriminator is provided with corresponding T1 images and text features as additional reference information. This design enables the discriminator to reference all available modalities when determining whether a PET image is real or synthetic. Specifically, the discriminator employs four convolutional layers to extract paired features from T1 and PET images, followed by merging these image features with text features in the discriminator fusion module.

The loss function comprises three primary components: the generator loss, the discriminator loss, and the Mean Squared Error (MSE) loss. The generator loss ($L_G$) is used during the training of generator and is defined as the negative expected value of the log-probability that discriminator assigns to synthetic PET images (where a probability closer to 1 indicates a more realistic image).
\begin{equation}
L_G(G, D) = -\mathbb{E}[\log(D(G(x_1, x_2),x_1,x_2))].
\end{equation}
$\mathbb{E}(.)$ denotes the operator of expected value calculation (averaging). $ G(x_1, x_2) $ denotes the PET image synthesized by the generator $ G $ from T1 image $ x_1 $ and text feature $ x_2 $. $ D(G(x_1, x_2), x_1, x_2) $ represents the probability assigned by the discriminator $D$ to the synthetic PET being real. Therefore, minimizing $L_G$ means maximizing this probability for all cases. After training, the generator is able to synthesize PET images that can deceive the discriminator.

During the training of the generator, the MSE loss ($L_{MSE}$) is also used. The MSE loss focuses on the voxel-level similarity between the synthetic and real PET images, complementing the generator loss by addressing details that may be overlooked. $ L_{MSE} $ calculates the expected value of the squared difference between corresponding voxels in the synthetic and real PET images.

\begin{equation}
L_{MSE}(G) = \mathbb{E}[(y - G(x_1, x_2))^2],
\end{equation}
where $ y $ denotes the real PET image.

The total generator loss combines $ L_G $ and $ L_{MSE} $, weighted by a parameter $ \lambda $:
\begin{equation}
L_{G_{\text{total}}}(G) = L_G(G, D) + \lambda L_{MSE}(G).
\end{equation}

During the training of the discriminator, the discriminator loss ($L_D$) is utilized. The discriminator's role is to distinguish between real and synthetic PET images. To achieve this, the discriminator loss is defined to maximize the probability that the discriminator assigns to real PET images (maximizing $D(y,x_1,x_2)$) and to minimize the probability assigned to synthetic PET images (maximizing $1 - D(G(x_1,x_2),x_1,x_2)$). Consequently, the discriminator loss is expressed as

\begin{equation}
L_D(G, D) = - \mathbb{E}[\log D(y,x_1,x_2)] - \mathbb{E}[\log(1 - D(G(x_1, x_2),x_1,x_2))].
\end{equation}

After training, the discriminator is able to distinguish between real and synthetic PET images, engaging in adversarial training with the generator.

\subsubsection{Fusion modules}

In Figure \ref{fig1b}, we design two fusion modules to integrate text and image features for the generator and discriminator respectively. The discriminator fusion module employs concatenation as previous generative models \cite{reed2016generative, zhang2018stackgan++}. This module applies a linear layer to adjust the dimensions of the text features from 756 to 32. These adjusted text features are then repeated to match the dimensions of the image features. Finally, the text and image features are concatenated along the channel dimension. Incorporating text features into the discriminator allows it to utilize these features when assessing the authenticity of PET images, thereby directing the generator to produce PET images that more accurately reflect the text information.

In the generator, we aim to use T1 features as the primary component, with text features serving to modulate them. Consequently, we designed the generator fusion module based on the idea of feature-wise linear modulation \cite{perez2018film}, rather than employing a concatenation. This module calculates a linear modulation based on the text feature to manipulate the entire image feature, resulting in a fused feature representation. Specifically, we employ two layers of MLPs to estimate scaling parameters (γ) and shifting parameters (β), which match the channel dimensions of the image features. These parameters are used to scale and shift the image features in each channel, facilitating the modulation of image features by the text features.

\subsection{Evaluation of synthetic PET images}

The PET images synthesized by our model are systematically evaluated from three key perspectives: image quality, diagnostic feature consistency, and clinical applicability. First, image quality is assessed based on the voxel-level and region-level visual resemblance of synthetic PET images to real PET images. Next, diagnostic feature consistency is examined to determine whether the diagnostic information in the synthetic PET images aligns with that in real PET images. Finally, clinical applicability is assessed to evaluate how synthetic PET images can enhance AD assessment in terms of diagnostic performance. 

\subsubsection{Evaluation of image quality}

The quality of synthetic PET images is evaluated from two complementary perspectives. On the one hand, we consider global image quality metrics that are widely used in computer vision. These voxel-level measures focus on the overall similarity between synthetic and real images. Specifically, we calculate the Structural Similarity Index (SSIM), Peak Signal-to-Noise Ratio (PSNR), and Mean Squared Error (MSE) \cite{hore2010image}. MSE quantifies the average voxel-wise discrepancy between synthetic and real PET images, PSNR represents the ratio between the meaningful signal and the error, and SSIM jointly assesses similarity in terms of luminance, contrast, and structure. While these global metrics are important for benchmarking generative models, they do not fully capture the clinical utility of PET synthesis.

On the other hand, because PET interpretation is usually performed at the brain region level, we further design region-based evaluation metrics tailored to this application scenario. For each subject, SUVRs are first averaged within anatomically defined regions, including 116 grey-matter regions from the AAL-116 atlas and four additional white-matter regions segmented using the SynthSeg toolbox \cite{billot2023synthseg}. Based on these regional SUVRs, we adopt two complementary measures. First, to assess consistency across regions within each subject, we compute the Pearson correlation coefficient (Pearson’s R) between synthetic and real PET regional SUVRs. This relationship is also visualized using scatter plots with fitted regression lines, providing an intuitive view of inter-region agreement. Second, to examine the distribution of synthesis errors across the brain, we calculate the absolute difference between synthetic and real PET SUVRs for each region and then average these errors across all subjects. This region-level error profile highlights anatomical areas where synthesis is more or less accurate and enables direct comparison between different methods.

In the results, our method is labeled as "T1+BBMs-LLM", which synthesizes PET images using T1-weighted images and BBMs encoded by LLM. We compare the proposed method to two representative baselines: (i) a classic image-to-image generation model using only T1 as input, labeled as "T1-only", and (ii) a model that incorporates BBMs as numerical values, rather than textual data, labeled "T1+BBMs-Num". In the "T1+BBMs-Num" method, BBMs are encoded into normalized numerical values, which are then directly used as features to assist in PET image synthesis. 

\subsubsection{Evaluation of diagnostic consistency}

\begin{figure}[H]
\centerline{\includegraphics[width=0.9\textwidth]{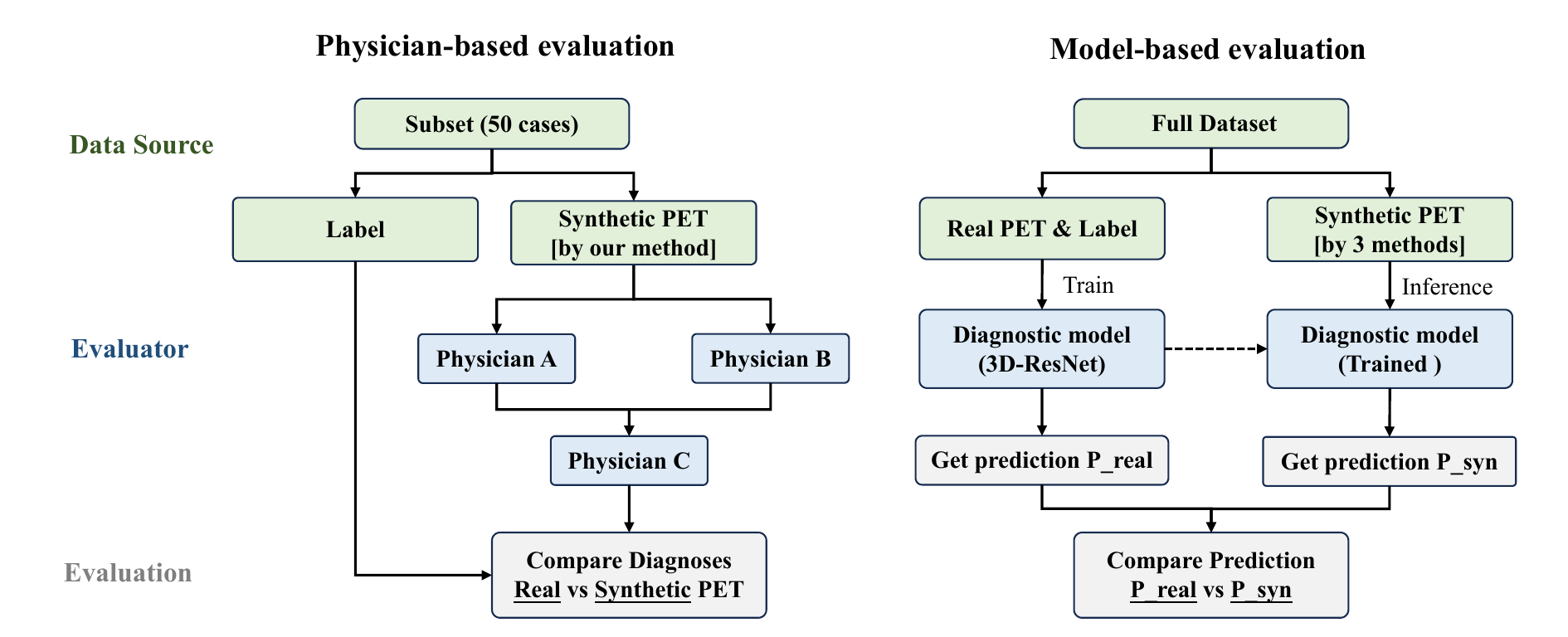}}
\caption{Overview of the diagnostic consistency evaluation framework. The physician-based evaluation uses a representative subset and a multi-reader arbitration workflow, while the model-based evaluation scales the assessment to the full dataset using a diagnostic model trained on real PET images.}
% }
\label{figeva}
\end{figure}

To determine whether synthetic PET images preserve diagnostically meaningful features, we designed a two-part evaluation framework combining expert physician review and a scalable model-based assessment. This framework aims to assess whether Aβ positivity can be reliably evaluated from synthetic PET images using the same clinical criteria applied to real PET images. A schematic summary of the complete diagnostic consistency evaluation workflow is shown in Fig.~\ref{figeva}. The evaluation includes (1) a physician-based assessment focusing on a representative subset and (2) a model-based assessment applied to the full dataset. Details of each component are described below.

Expert interpretation of PET imaging requires substantial clinical experience and time; therefore, this evaluation was performed on a carefully selected subset of 50 cases. To ensure representativeness, the subset was chosen to match the MSE distribution of the full dataset (see Supplementary Table S1), avoiding bias toward atypical cases. For each selected case, two senior physicians independently assessed Aβ positivity based solely on the synthetic PET images generated by our method, blinded to all clinical information and the reference label. Any disagreement was adjudicated by a third senior physician. The adjudicated diagnosis from synthetic PET was then compared against a fixed reference label established previously from the real PET together with relevant clinical information under routine clinical procedures. This double-reading with arbitration mirrors standard clinical practice and provides an unbiased human benchmark for diagnostic consistency.

To scale the evaluation beyond the physician-reviewed subset and to enable direct comparison across different synthesis methods, we complemented the expert assessment with a model-based evaluation. A 3D-ResNet~\cite{chen2019med3d} classifier was trained on real PET images and their clinical labels to learn diagnostic patterns consistent with physician decision-making. After validation on real PET data, the trained model was applied to synthetic PET images generated by all three synthesis pipelines, including our proposed method, the “T1-only’’ method, and the “T1+BBMs-Num’’ method. By comparing the model’s predictions on real PET (P\textsubscript{real}) with its predictions on synthetic PET produced by each method (P\textsubscript{syn}), we assess not only whether diagnostic features are preserved within our method but also how its diagnostic consistency compares with alternative synthesis strategies. Although the results of this model-based evaluation are influenced by the classifier’s own performance, the model demonstrated sufficiently stable accuracy on real PET and therefore provides a reliable auxiliary tool for large-scale consistency assessment and cross-method comparison, without substituting clinical judgment.

Diagnostic consistency between real and synthetic PET–based assessments was quantified using Cohen’s Kappa~\cite{cohen1960coefficient}. Kappa values between 0 and 0.2 indicate poor consistency, 0.2 to 0.6 indicate moderate consistency, and values $\ge$0.6 indicate good consistency. Because Aβ positivity prediction is a binary classification task, we additionally report accuracy, sensitivity, specificity, and F1-score.

\subsubsection{Evaluation of clinical applicability of fully AI pipeline}

To further explore and evaluate the potential of synthetic PET images, we develop an AI diagnostic model based on synthetic images and assess the model performance on the clinical applicability.

In the absence of real PET images, we propose two models to evaluate the utility and scalability of synthetic PET images. The first model relies solely on synthetic PET images, while the second model integrates synthetic PET images with all available clinical assessments to enhance diagnostic performance. The first model is trained based on the 3D-ResNet architecture. The second model employs the well-trained first model for feature extraction from synthetic PET images and the established MLP model (same as described in the LLM-based text encoder section) for feature extraction from clinical assessments. These features are then fused at a logits layer.

We compare these two proposed models with models based on (i) T1, (ii) BBMs, and (iii) a combination of T1 and BBMs. This comparison aims to determine whether synthetic PET images offer advantages in predicting Aβ positivity and whether they can be further optimized by incorporating additional relevant variables. The evaluation metrics include AUC, accuracy, sensitivity, specificity, and F1 score.

\subsection{Implementation}

For the generative model, the loss function weight $\lambda$ is set to 10 to balance the losses. The Adam optimizer is used with an initial learning rate of 5.0e-4 for the generator and 1.0e-4 for the discriminator. The weight decay is set to 1.0e-4, and the parameters for the first and second moment estimates of the gradients, β1 and β2, are set to 0.5 and 0.999, respectively. The learning rate decay strategy employs cosine annealing with a period of 100 epochs and a minimum learning rate of 1.0e-7. Training is conducted for 100 epochs with a batch size of 8. 

For the classification model, the initial learning rate is set to 1.0e-4, with the same optimization strategy and number of epochs as the generative model, but with a batch size of 16. All models are initialized using Gaussian initialization (mean = 0, standard deviation = 0.02). Training is implemented using the PyTorch framework and performed on a single NVIDIA GeForce GTX 3090Ti GPU.

\subsection{Statistical analysis}

Due to the non-normal distribution of the data, differences in gender are tested using Fisher's exact test, while differences in age, education years, BBMs, and NAs are tested using the two-sided Mann–Whitney U test. For model validation, the mean and standard deviation (STD) of performance metrics from cross-validation are computed and reported. Performance metric comparisons between different methods are conducted using one-sided paired t-test. All statistical analyses are performed using python 3.10, scikit-learn 1.5.0 and scipy 1.13.1.

\section{RESULTS}\label{sec3}

\begin{table}[htbp]
\centering
\begin{threeparttable}

% ==== 表题（10pt, Times New Roman, Bold）====
\caption{\fontsize{10pt}{12pt}\selectfont Participant characteristics}
\label{table1}

% ==== 表格本体（8pt, Times New Roman, 单倍行距）====
{\fontsize{8pt}{10pt}\selectfont
\begin{tabularx}{\textwidth}{p{5cm} X X X}
\hline
\textbf{} & \textbf{CU} & \textbf{MCI} & \textbf{ADD} \\  % 表头加粗
\hline
Sample Size & 342 & 134 & 90 \\
Age (median (IQR)) & 65.0 (10.0) & 66.0 (10.0) & 66.0 (11.0) \\
Females (No. (\%)) & 218 (63.7) & 83 (61.9) & 52 (57.8) \\
Education (median (IQR)) & 12.0 (5.0) b*** & 11.0 (4.0) a*** & 9.0 (6.0) c*** \\
A$\beta$40 (pg/mL) (median (IQR)) & 196.43 (65.3) & 193.05 (81.9) & 197.40 (59.6) \\
A$\beta$42 (pg/mL) (median (IQR)) & 10.2 (4.13) b** & 10.3 (4.00) & 9.02 (4.26) c** \\
A$\beta$42/40 rate (median (IQR)) & 0.05 (0.02) b*** & 0.05 (0.01) & 0.04 (0.01) c*** \\
T-Tau (pg/mL) (median (IQR)) & 2.20 (1.19) & 2.41 (1.28) & 2.18 (1.55) \\
P-Tau 181 (pg/mL) (median (IQR)) & 1.75 (1.06) b*** & 1.97 (1.56) a* & 3.36 (2.74) c*** \\
NFL (pg/mL) (median (IQR)) & 12.6 (8.25) b*** & 14.8 (9.16) a*** & 19.6 (9.18) c*** \\
P-Tau 181/A$\beta$42 rate (median (IQR)) & 0.18 (0.12) b*** & 0.19 (0.17) & 0.4 (0.37) c*** \\
MMSE (median (IQR)) & 28.0 (2.0) b*** & 27.0 (2.0) a*** & 18.0 (6.0) c*** \\
MoCA-B (median (IQR)) & 26.0 (4.0) b*** & 22.0 (4.0) a*** & 14.0 (6.0) c*** \\
AVLT-LDR (median (IQR)) & 5.0 (4.0) b*** & 2.0 (3.0) a*** & 0.0 (1.0) c*** \\
AVLT-R (median (IQR)) & 22.0 (2.0) b*** & 18.0 (3.0) a*** & 17.0 (4.5) c** \\
AFT (median (IQR)) & 16.0 (5.0) b*** & 13.0 (4.0) a*** & 9.0 (5.0) c*** \\
BNT (median (IQR)) & 25.0 (5.0) b*** & 21.0 (5.0) a*** & 19.0 (7.0) c*** \\
STT-A (median (IQR)) & 44.0 (16.0) b*** & 51.0 (19.0) a*** & 75.0 (37.8) c*** \\
STT-B (median (IQR)) & 117.0 (45.0) b*** & 147.0 (53.2) a*** & 190.0 (84.0) c*** \\
A$\beta$-PET positivity (No. (\%)) & 110 (32.2) b*** & 54 (40.3) & 72 (80.0) c*** \\
\hline
\end{tabularx}
}
% ==== 表注（10pt, 单倍行距）====
\begin{tablenotes}
\fontsize{10pt}{12pt}\selectfont
\item \textbf{Note:} Data are presented as median (M) and interquartile range (IQR) or participant number (n) and percentage (\%). 
Data were compared using a two-tailed Mann–Whitney U test or Fisher’s exact test.
\item \textbf{Abbreviations:} A$\beta$, amyloid-$\beta$; NFL, neurofilament light chain; IQR, interquartile range; 
MMSE, Mini-Mental State Examination; MoCA-B, Montreal Cognitive Assessment–Basic Version; AVLT, Auditory Verbal Learning Test; 
AVLT-LDR, delayed free recall of the Auditory Verbal Learning Test; BNT, Boston Naming Test; AFT, Animal Fluency Test; STT, Shape Trails Test.
\item \textbf{Significance:} (a) Significant difference between CU and MCI; (b) between CU and ADD; (c) between MCI and ADD. 
*, **, and *** indicate significant differences between two groups after Bonferroni correction, with p < 0.05, p < 0.01, and p < 0.001, respectively.
\end{tablenotes}

\end{threeparttable}

\end{table}

\subsection{Participant characteristics}

This study includes 566 participants with three groups: cognitively unimpaired (CU, n = 342), mild cognitive impairment (MCI, n = 134), and AD dementia (ADD, n = 90). The characteristics of these participants are depicted in Table \ref{table1}. The age and sex distributions are similar across the groups. The median education level is 12.0 years in the CU group, which is significantly higher than the 11.0 years in both the MCI and ADD groups. Furthermore, the education level in the MCI group is significantly higher than that in the ADD group. Regarding plasma biomarkers, the ADD group exhibited significantly higher levels of p-tau181, NFL, and P-Tau181/Aβ42 ratio, as well as markedly lower levels of Aβ42 and Aβ42/40 ratio compared to the CU and MCI groups. The MCI group also shows significant alterations in p-tau181 and NFL biomarker levels compared to the CU group, although to a lesser degree than the ADD group. Cognitive and functional assessments, including MMSE, MoCA-B, and other 6 NAs (AVLT-LDR, AVLT-recognition, AFT, and BNT) in cognitive sub-domains, demonstrate a clear decline from CU to MCI and ADD. Conversely, STT-A and STT-B exhibit a significant increasing trend across all three groups.

\subsection{Evaluation of image quality}

We evaluate the quality of the synthetic Aβ-PET images by assessing both the overall image fidelity and accuracy in different brain regions. Our method (T1+BBMs-LLM) demonstrates superior performance compared to other methods in these aspects. 

\begin{figure}[H]
\centerline{\includegraphics[width=0.9\textwidth]{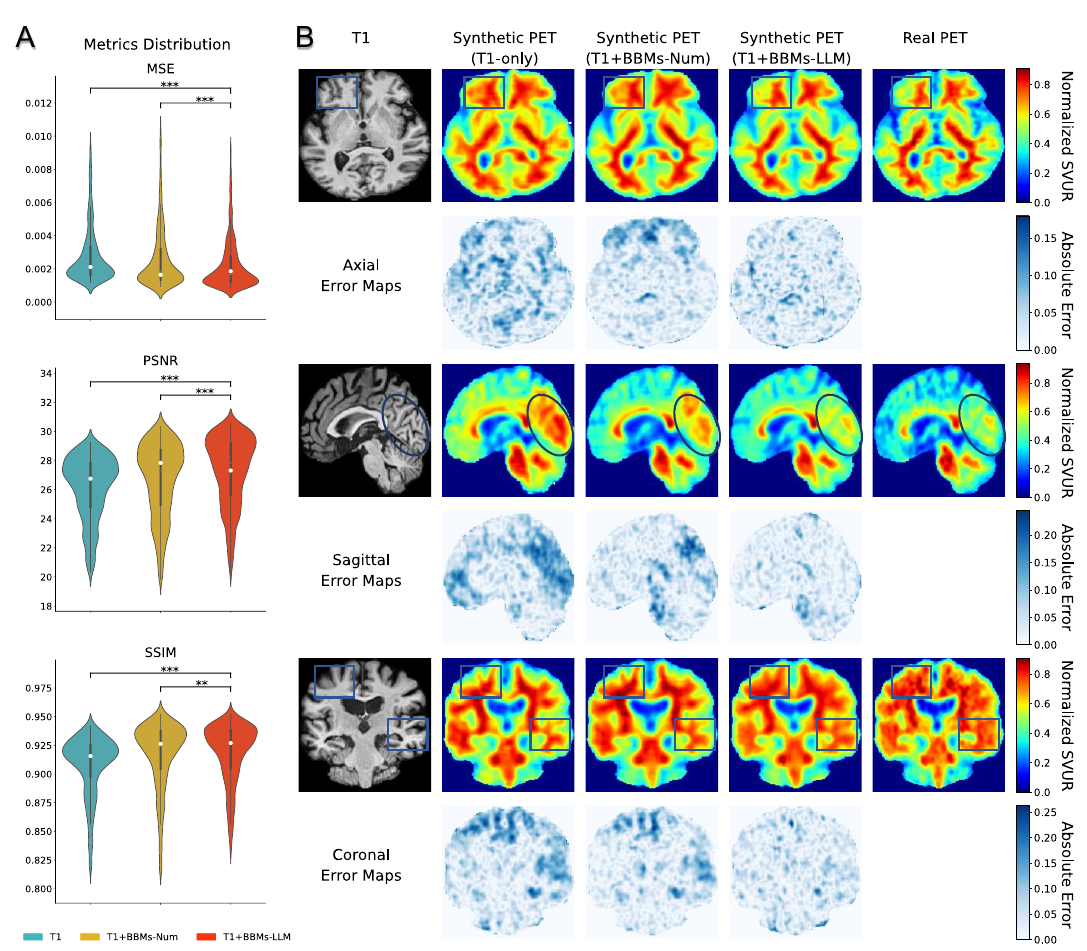}}
\caption{Evaluation of image quality of the synthetic PET image. (A). Violin plots for comparing image quality assessment metrics distributions in individual predictions across methods. **: indicates significance at P < 0.01. ***: P < 0.001. (B) Visual comparison for investigating the methodological differences through case studies. From left to right, it includes T1, synthetic PET images from three methods, and real PET images. PET images are displayed using range-normalized SUVRs with pseudocolors for clarity. Areas with obvious differences are highlighted with annotations, as boxes or circles, in the images. Below each synthetic PET image, corresponding error maps are provided, indicating discrepancies measured by absolute differences. 
}
\label{fig3}
\end{figure}

As shown in Figure~\ref{fig3}A, our method achieves the lowest MSE of 0.00235±0.00021 and the highest PSNR of 27.24±0.26 dB, indicating that the synthesized PET images have minimal voxel-level discrepancies from real PET images. In addition, an SSIM of 0.9202±0.0033 further confirms that our method preserves overall structural similarity. The violin plots also show that the distribution of evaluation results is more favorable for our method. Statistical tests support these findings, with our method showing significant improvements across all metrics (MSE: P = 1.85e-14 for T1-only and P = 4.47e-6 for T1+BBMs-Num; PSNR: P = 2.46e-37 for T1-only and P = 2.90e-11 for T1+BBMs-Num; SSIM: P = 7.92e-67 for T1-only and P = 0.0067 for T1+BBMs-Num). Notably, the T1+BBMs-Num method shows improvement over the T1-only method, underscoring the usefulness of incorporating BBMs in synthesizing PET images. However, this approach is limited by numerical encoding. Our method overcomes this limitation by introducing LLM-based encoding, thereby better leveraging BBMs.

Qualitative comparisons are presented in Figure~\ref{fig3}B. Each case includes T1 images, real PET images, synthetic PET images generated by three methods, and corresponding absolute error maps. These cases feature coronal, sagittal, and transverse slices for a comprehensive illustration. PET images synthesized from T1-only inputs mainly preserve structural information but fail to capture pathological uptake patterns, leading to larger errors. The T1+BBMs-Num method improves the representation of Aβ deposition but still shows noticeable deficiencies in regions such as the frontal, occipital, and temporal lobes (highlighted with boxes and circles). In contrast, our method produces synthetic PET images with spatial patterns closely resembling real PET, reducing both local discrepancies and global error.

\begin{figure}[H]
\centerline{\includegraphics[width=0.9\textwidth]{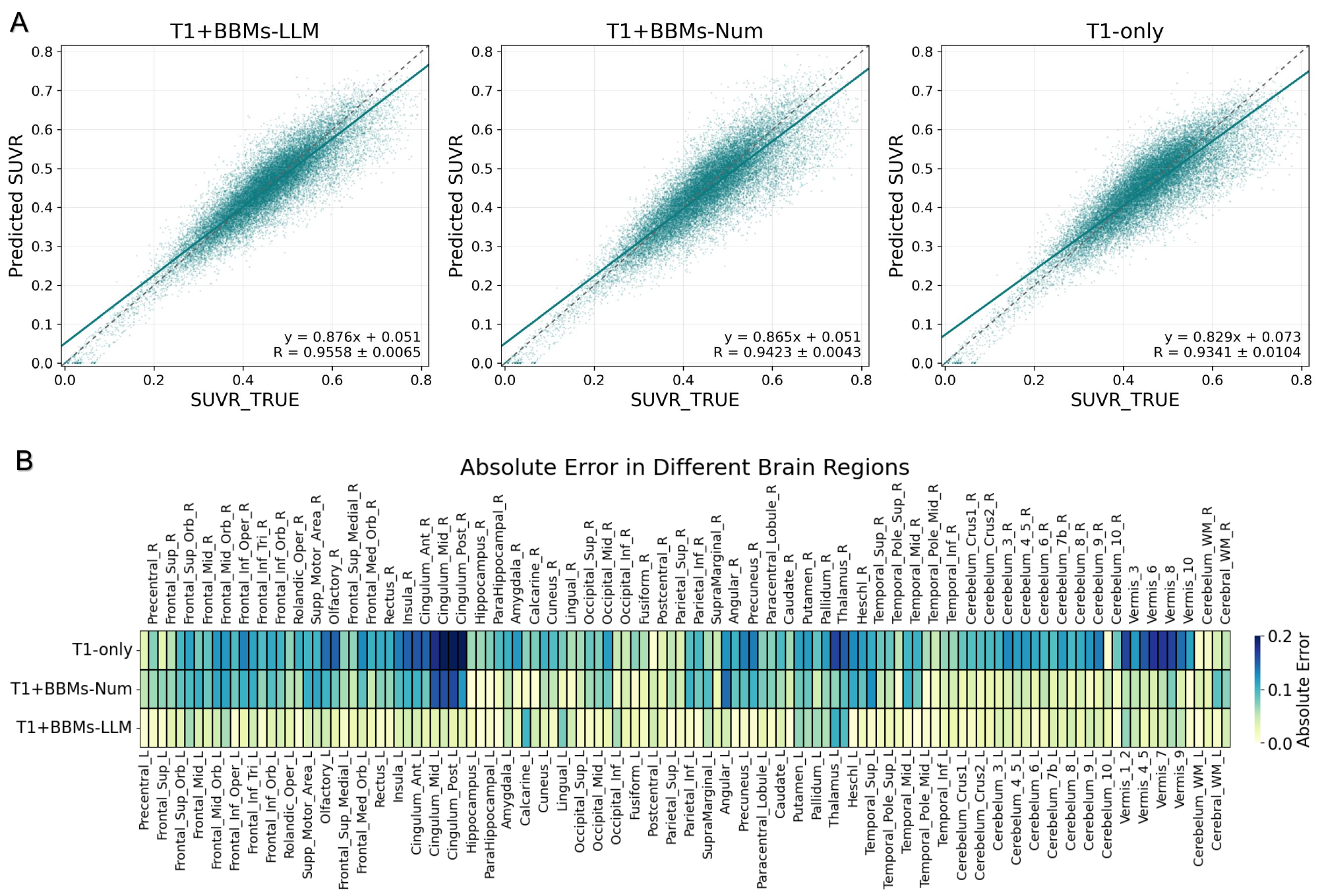}}
\caption{Region-based evaluation metrics for different methods. (A). Scatter plot between ground-truth and predicted regional SUVRs for all subject with inter-region correlations and (B) absolute errors in different brain regions. The brain is divided into 116 grey-matter regions based on the AAL-116 atlas and 4 white-matter regions. The correlation and absolute error between the mean SUVR of the synthetic PET and real PET images for each region are shown. In B, darker colors representing larger errors. Each row corresponds to a method, with the last row representing our method.
}
\label{fig3s}
\end{figure}

To further evaluate region-specific accuracy, we compared synthetic and real PET SUVRs at the level of anatomically defined regions. Our method achieves the strongest correlation, with Pearson’s R of 0.9545±0.0065, demonstrating robust agreement in regional uptake patterns (Figure~\ref{fig3s}A). Comparison of individual R among three methods suggest a significant improvement of our method (Pearson's R: P = 2.42e-42 for T1-only and P = 5.20e-18 for T1+BBMs-Num). Figure~\ref{fig3s}B shows the distribution of regional absolute errors across 116 grey-matter and 4 white-matter regions, averaged over all subjects. Our method consistently yields lower errors than the baselines, particularly in regions such as the frontal cortex, temporal lobe, and precuneus—areas known to be clinically relevant for Aβ deposition. These results further confirm that our method not only improves global similarity but also better captures the spatial distribution of pathological features across the brain.

Figure~\ref{fig3s}B presents the distribution of regional absolute errors across 116 grey-matter and 4 white-matter regions, averaged over all subjects. Our method provides the lowest discrepancy in regional SUVRs compared with the baselines, highlighting its ability to accurately capture regional A$\beta$ deposition patterns. Notably, the bilateral cingulum (Cingulum\_Mid\_L and Cingulum\_Mid\_R), cerebellar vermis (Vermis\_3 and Vermis\_4\_5), and bilateral thalamus exhibit markedly reduced errors, while consistent improvements are also observed in subcortical structures such as the caudate, putamen, pallidum, and insula bilaterally. These results confirm that our method not only improves global similarity but also more faithfully preserves the spatial distribution of pathological features across the brain.

\subsection{Evaluation of diagnostic consistency}

\begin{figure}[H]
\centerline{\includegraphics[width=0.9\textwidth]{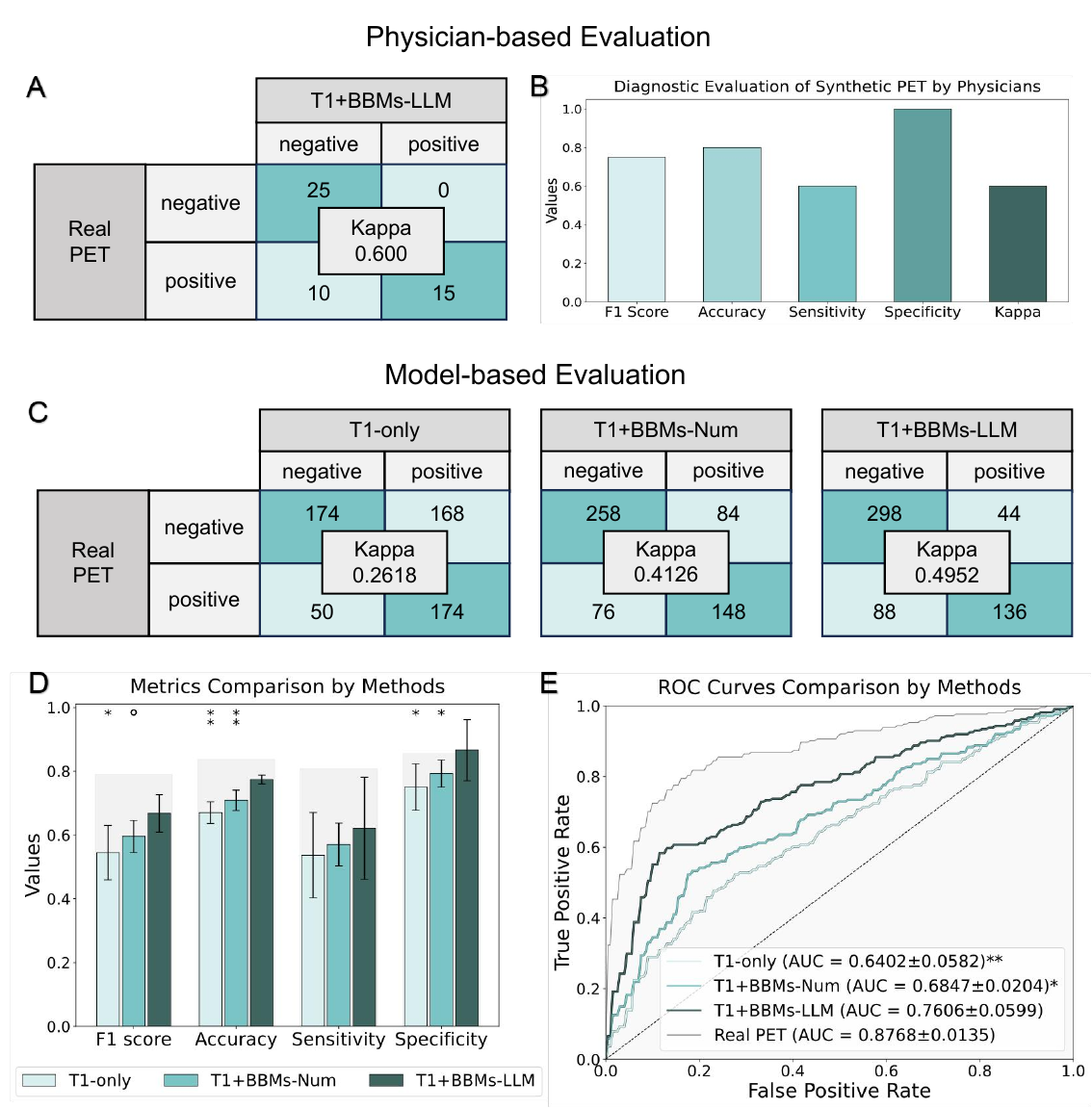}}
\caption{Evaluation on diagnostic consistency of the synthetic PET image. (A-B) The corresponding results of physician-based evaluation. Only results from our method participant this analysis. (A) Inter-rater agreement matrices for real PET and the corresponding synthetic PET from our method, evaluated by physicians. (B) Bar charts of performance metrics of our synthetic PET in physician-based evaluation. (C-E) The performance of PET synthesized by three methods, evaluated by a model trained on real PET. (C) The inter-rater agreement matrix comparing diagnosis of real and synthetic PET, with Cohen's kappa in the central box. (D) Bar charts showing the F1 score, accuracy, and sensitivity of the three methods, represented by their means with error bars indicating standard deviations. (E) Receiver operating characteristic (ROC) curves and the corresponding AUCs. The performance of real PET is shown with a grey border as a reference. *: indicates significance at P < 0.05. **: P < 0.01. $\circ$: a marginal significance, where the P-value is 0.0532.}
% }
\label{fig4}
\end{figure}

To examine whether synthetic PET images retain diagnostically meaningful features beyond visual similarity, we conducted a physician-based evaluation. As shown in Figure~\ref{fig4}A–B, the synthetic PET images achieved an accuracy of 0.80, an F1 score of 0.75, and a Cohen’s Kappa of 0.60. These values indicate good agreement with diagnoses based on real PET images, suggesting that the synthetic images preserve critical pathological features relevant for clinical interpretation. Notably, diagnostic consistency was higher for Aβ-negative cases, with relatively lower performance observed for Aβ-positive cases. This discrepancy likely stems from the greater challenge faced by generative models in accurately capturing and reproducing the subtle pathological signals characteristic of Aβ-positive subjects.

% Additionally, we perform a model-based evaluation using all data samples (Figure Sx). The analysis suggests that our method outperforms T1-only and T1+BBMs-Num methods in terms of diagnostic consistency, supported by higher inter-rater agreements and diagnostic performance metrics from the judgment model. 

For the model-based evaluation, the judge model trained on real PET images demonstrates strong performance (accuracy of 0.8357 and an AUC of 0.8768). We see that the judge model captures key diagnostic features in real PET images and is thus well suited for evaluating the diagnostic consistency. Although this result may be influenced by the imperfect accuracy of judge model, it allows for a comparative analysis with our approach evidently outperforming the other two baselines.

% After inference with the judge model, the diagnostic results for both real and synthetic PET images are obtained. The agreement matrices for these results are shown in Figure \ref{fig4} C. After inference with the judge model, the diagnostic results for both real and synthetic PET images are obtained. 

Figures \ref{fig4}C-E further illustrate the diagnostic results inferred by the judge model. We recognize that the T1-only method performs poorly across all metrics, with an F1 score and sensitivity below 0.7, and T1+BBMs-Num performs slightly better (merely above 0.7). These results indicate that the PET images synthesized by these methods remarkably differ from real PET images in terms of the diagnostic features. In contrast, our method achieves the leading values in all metrics, reflecting the strength of our diagnostic consistency to the real PET. For instance, our method outperforms the two compared methods in accuracy (P = 0.0015 for T1-only and P = 0.0060 for T1+BBMs-LLM ) and AUC (P = 0.0021 for T1-only and P = 0.0236 for T1+BBMs-LLM ). Considering the class imbalance, the F1 score is another important evaluation metric. Our method also shows an improvement over the competing methods (P = 0.0135 for T1-only and P = 0.0532 for T1+BBMs-Num). The agreement matrices for these results are shown in Figure \ref{fig4} E. Our method achieves the highest Cohen’s Kappa among all competing methods. Note that the Cohen’s Kappa in this analysis is slightly lower than that using human-based evaluation, which is induced by the imperfect diagnosis ability of the judge model. Overall, though being partially biased, the model-based evaluation offers reasonable evidence supporting the advantage of our method in terms of consistent diagnostic feature generation, which is in line with the image quality metrics analysis.

\subsection{Evaluation of clinical applicability of fully AI pipeline}

\begin{figure}[H]
\centerline{\includegraphics[width=1\textwidth]{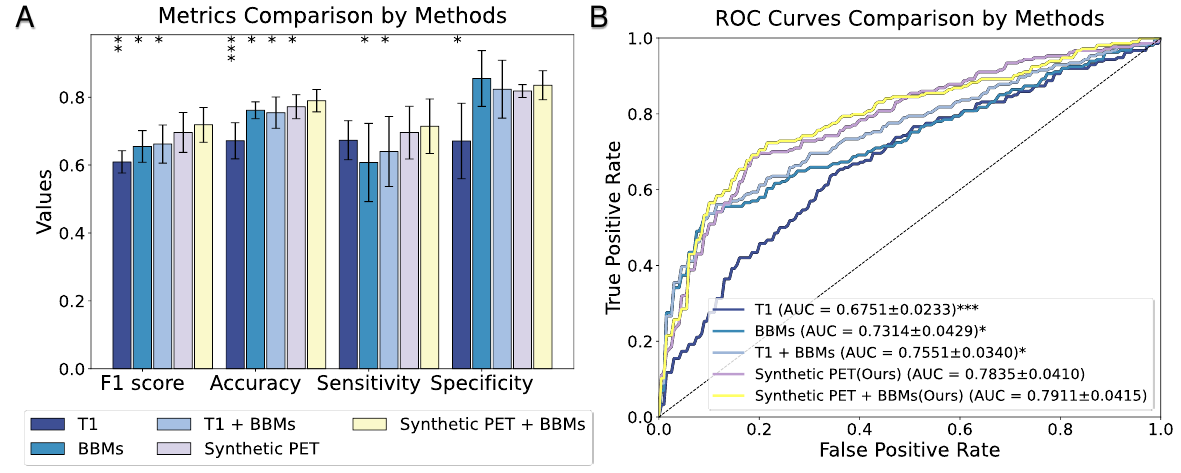}}
\caption{Evaluation results on clinical applicability of the synthetic PET image. (A) Bar charts of F1 score, accuracy, and sensitivity, with error bars indicating standard deviations. (B) ROC curves with the corresponding AUCs. For both plots, the significant differences are annotated for the fusion of synthetic PET and BBMs compared to other methods. *: indicates significance at P < 0.05. **: P < 0.01. ***: P < 0.001.}
\label{fig5}
\end{figure}

We utilize the synthetic PET images to train Aβ pathology classification models, establishing a complete AI pipeline to further facilitate the clinical applicability of PET generation. We use T1 images and BBMs to diagnose Aβ positivity, as comparison baselines. As shown in Figure \ref{fig5}, T1 achieves a classification AUC of only 0.64, with other metrics also performing poorly. BBMs yields slightly better results but with a higher rate of false negatives. Combining T1 images and BBMs using a multimodal diagnostic model improves all metrics, yet this approach still falls short compared to our synthetic PET images. This finding indicates that integrating T1 and BBMs through PET generation model offers a more cohesive fusion for AD diagnosis compared to traditional multimodal diagnostic models. Furthermore, fusing synthetic PET with BBMs yields superior results that outperforms single-modal approaches or their simple fusion (T1, BBMs, and T1+BBMs). Also, the model integrating synthetic PET and BBMs demonstrates a balanced sensitivity and specificity. These findings underscore the potential of synthetic PET from our proposed method in AD diagnostics, particularly in enhancing pathological diagnosis accuracy by integrating diverse sources of clinical information. 

% Note that the synthetic PET is generated using the initial diagnosis from the same BBM-based model for prompting (see Methods). This observation indicates that the final property of synthetic PET is not fully determined by the initial diagnosis from BBMs. 

\subsection{The effectiveness of prompt engineering}

\begin{figure}[H]
\centerline{\includegraphics[width=1\textwidth]{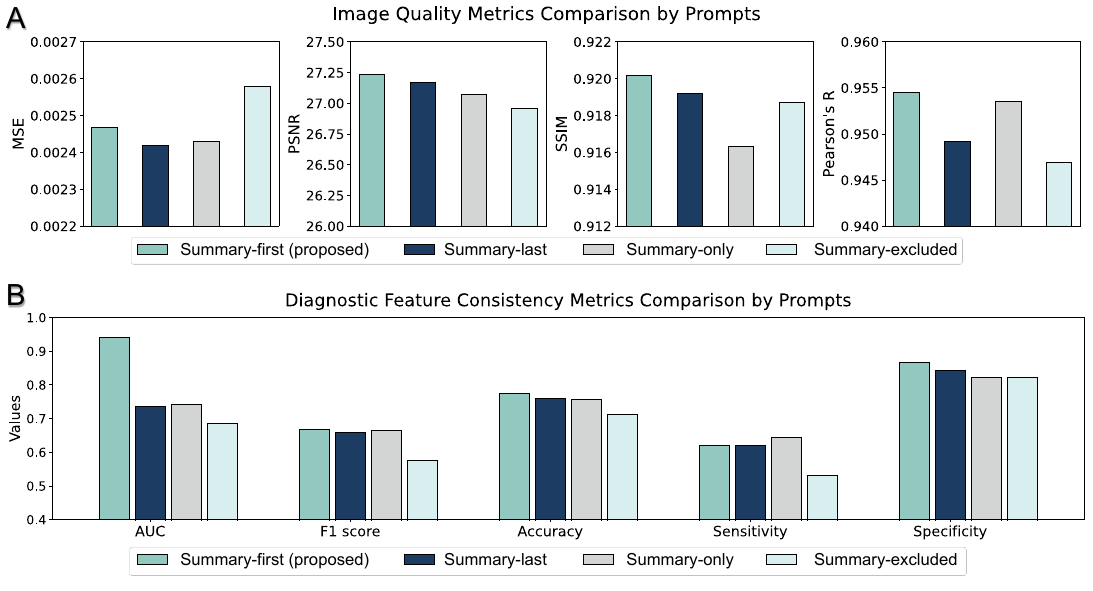}}
\caption{Ablation study for the effectiveness of prompt engineering. Evaluation of (A) PET image quality and (B) diagnostic feature consistency using our proposed prompt and three different prompts for comparisons. The results in B is obtained using model-based feature consistency evaluation. "Summary-first" indicates our proposed design. "Summary-last" places the summary sentence at the end. "Summary-only" uses only the summary sentence. "Summary-excluded" does not include the summary sentence and preserves all other prompts about the clinical variables.}
\label{fig6}
\end{figure}

The prompt engineering for utilizing multi-modality clinical data (e.g., demographics, BBMs and NAs) is a core contribution in our study. We demonstrate the effectiveness of the proposed prompt design using a comparison to three variants in the prompt. In Figure \ref{fig6} A and B, we observe a remarkable performance decrease when placing the summary sentence (the diagnosis/prediction prompt in Figure 1A) at the end of the prompt. Also, excluding the diagnostic sentence degrades all metrics. However, using the summary sentence only does not result in leading performance, which suggests the key descriptions about the clinical variables are informative. Overall, we recognize that using the summary sentence in starting a prompt can efficiently guide the LLM to encode clinical variables and enhance the quality of synthetic PET images.

\section{Discussion}\label{sec4}

In this study, we propose a multimodal generative AI approach for synthesizing PET images from structural MRI and BBMs. We develop a language-enhanced framework and prompt engineering to optimize PET synthesis. Our systematic validation shows that the proposed method can (i) produce synthesized PET semantically similar to real-world PET at individual level; (ii) offer consistent Aβ pathological features to support clinical assessment; and (iii) enable the development of fully AI pipeline to provide Aβ diagnosis without requiring real-world PET scans. 

Achieving a cost-effective and accurate AD diagnosis remains a significant challenge. Routine diagnosis using Aβ-PET scans is heavily constrained by the cost, availability, and risk of radiation exposure \cite{rice2017diagnostic,nordberg2010use}. Meanwhile, BBM-based methods provide a more accessible alternative for predicting AD positivity without requiring PET scans, but the lack of spatial information in BBMs limits their ability to fully replace PET imaging \cite{devanarayan2024plasma, brum2024blood}. Our multimodal framework integrates pathological insights from BBM features and T1-based anatomical details to synthesize high-quality Aβ-PET images. Our validation demonstrates that these images not only provide the spatial information necessary for accurate AD assessments, but also enable the development of AD diagnostic models that outperform the BBM-based methods. Additionally, integrating synthetic Aβ-PET images with BBMs further enhances diagnostic accuracy, highlighting the potential of generative models to improve AD diagnosis. Overall, our study opens avenues to synthesize PET images from MRI data and BBMs towards predictive AD diagnosis. This data-driven approach is extensible to large-scale early AD diagnosis by offering a cost-effective alternative to real PET imaging.

Our approach is pioneering in integrating BBMs into a multimodal generative model for PET synthesis. We show that the BBM-detected pathological insights can enhance the synthesis of Aβ-PET spatial pattern, complementing to traditional MRI-based generative approaches. We advance the representation of BBMs by employing a language-enhanced encoder, thereby improving the integration of non-imaging data with T1 images and enhancing PET synthesis.
Conventionally, BBMs and other clinical variables were encoded as numerical formats (e.g. concentration levels of BBMs) \cite{xia2021learning, xia2019consistent}. A simple integration of high-dimensional image features (more than 10 thousands) and low dimensional non-image BBM features (represented as a few numbers) would likely results in a performance biased towards the image features \cite{cui2023deep}, which limits the contribution of non-imaging features. In contrast, our approach employs prompt engineering to map non-imaging data into meaningful contexts, guiding a medical LLM to extract knowledge-enhanced features. To improve the context understanding, we design a "summary-first" prompt that prioritizes the critical diagnostic information. Experimental results demonstrate that this design enhances the quality of synthetic PET images, highlighting that a specialized prompt design can greatly contribute to the overall performance.

A key contribution of our study is to enable a systematic process of PET images synthesis and clinical application evaluation. Our comprehensive evaluation is a progressive process, encompassing image quality, diagnostic consistency, and clinical applicability. By contrast, prior PET image synthesis studies have primarily focused on metrics of visual quality (e.g., MSE, PSNR, and SSIM) \cite{hu2021bidirectional, zhang2022bpgan, vega2024image}. It becomes increasingly evident that the detailed depiction of anatomical structures and image-based pathological characteristics are crucial in clinical decision making that are beyond the scope of traditional metrics. In our study, recognizing that PET analysis is frequently performed at the brain-regional level, we introduce key evaluation metrics based on regional SUVRs, including an absolute error and inter-region correlation. These metrics ensure a clinically-meaningful structural similarity between synthetic PET images and real PET images. Additionally, physician-based evaluation of image-based pathological characteristics can help assess the effectiveness of synthetic PET images under a clinical setting. However, a physician-based evaluation over large-scale data is time-consuming and labor-intensive. To respond, we perform a physician-based evaluation on a subset of the data and conducted model evaluation on the entire dataset. For the model evaluation, we develop an AI model that can mimic human diagnostic criteria and offer quality insights into the PET image generation. Finally, we establish a fully AI-driven pipeline, consisting of PET generation and AI diagnosis, to substantially confirm the clinical applicability of the proposed method.

% limitations and perspective 
Our study has several limitations. First, due to the difficulty in multimodal data acquisition, our study has not undergone a multi-center systematic evaluation. In addition, the range of the BBM values could vary across the assays and reference protocols \cite{giangrande2023harmonization} and a quality assessment is required to reveal potential protocol-induced BBM variances. Moreover, integrative analysis with other promising biomarkers is valuable to extend the generalization power of our approach. For instance, P-Tau 217 is suggested to predict continuous brain Aβ levels in early AD subjects \cite{devanarayan2024plasma} and is recommended as the only BBM to diagnose Aβ pathology in the NIA-AA 2024 Diagnostic Guidelines \cite{jack2024revised}. Inclusion of P-Tau 217 into our framework could help achieve the performance improvement. Finally, considering the association of Tau-PET with T1 and BBM \cite{lee2024synthesizing, matthews2024relationships}, the extensive structure of our framework could be adapted for Tau-PET synthesis to facilitate rapid AD staging and prognosis.

In conclusion, our research presents a language-enhanced, multimodal framework that effectively synthesizes Aβ-PET images from T1 images and BBMs. This advancement addresses critical limitations of traditional PET imaging data, offering a cost-effective and accurate alternative that enhances early AD assessment, diagnosis, and decision making.

% TC:ignore
%\backmatter
\section*{Contributors}

ZZ and XM are major contributors to draft and revise of the manuscript for content and to analyse the data.
XM, QG, QH and FX played major role in the acquisition of data.
ZZ, XM, QG, SZ, QH, MZ, and ML substantially revise the manuscript. 
ZZ, QH, MZ, ML and FX conceptualize and design the study.
ZZ, XM, QH, MZ, and ML interpret the data. 

\section*{Declaration of interests}

All authors declare no potential conflict of interests.

\section*{Acknowledgments}
This study is supported in part by the National Natural Science Foundation of China (82402394, 82201583, 8217052097, 82071962), Shanghai Pujiang Program ( 23PJ1430200), STI2030-Major Projects (022ZD0213800), and Shanghai Artificial Intelligence Laboratory.

\section*{Data Sharing Statement}

Due to patient privacy protection, institutional regulations, and restrictions imposed by the ethical approvals governing this study, the raw PET, MRI, and clinical data cannot be made publicly available. De-identified derived data supporting the findings of this study can be provided upon reasonable request to the corresponding author and will require approval from the institutional review boards of Huashan Hospital and the Sixth People's Hospital affiliated with Shanghai Jiao Tong University.

\bibliographystyle{elsarticle-num} 
\bibliography{cas-refs}
%TC:endignore

\end{document}